\documentclass[11pt]{article}

\usepackage{acl}

\usepackage{times}
\usepackage{latexsym}

\usepackage[T1]{fontenc}

\usepackage[utf8]{inputenc}

\usepackage{microtype}

\usepackage{inconsolata}

\usepackage{graphicx}

\usepackage[size=tiny]{todonotes}

\title{Feedback as an Interface: Controlling Form and Detail in LLM Translation Refinement}

\title{Feedback as an Interface: Controlling Form and Detail in LLM Translation Refinement}

\author{
 \textbf{Yiqi Liu\textsuperscript{1}}, \quad
 \textbf{Flavjo Tomori\textsuperscript{2}}, \quad \\ 
 \textbf{Nafise Sadat Moosavi\textsuperscript{3}}, \quad 
 \textbf{Steffen Eger\textsuperscript{4}},  \quad
 \textbf{Chenghua Lin\textsuperscript{1}\thanks{Corresponding author.}}\vspace{0.2cm}
\\
 \textsuperscript{1}University of Manchester, \hspace{0.07cm}  
 \textsuperscript{2} Osapiens, \hspace{0.07cm}  \\
  \textsuperscript{3}University of Sheffield,  \hspace{0.07cm}
 \textsuperscript{4}University of Technology Nuremberg
\\
\texttt{\{yiqi.liu, chenghua.lin\}@manchester.ac.uk} \quad \texttt{flavjotomori@gmail.com}     \\
\texttt{n.s.moosavi@sheffield.ac.uk} \quad \texttt{steffen.eger@utn.de}\quad 
\\
}

\usepackage{booktabs}
\usepackage{multirow}
\usepackage{diagbox}
\usepackage[dvipsnames]{xcolor}
\usepackage{placeins}

\usepackage{array}
\usepackage[most]{tcolorbox}

\usepackage{amssymb}
\usepackage{makecell}

\begin{document}

\maketitle

\begin{abstract}
Feedback presentation shapes how LLMs use evaluator judgments during translation refinement. We introduce a machine translation protocol that keeps the evaluator score, main issue, and recommendation fixed while varying feedback form and supporting detail. Across seven models and six translation directions, low-load feedback outperforms high-load feedback in all 42 translator-direction settings under xCOMET. On the 29,400-record rewriting subset, mean xCOMET scores multiplied by 100 are 83.30 for low-load feedback and 80.67 for high-load feedback; the full-matrix original-translation reference mean is 83.46. MetricX-24 supports the low-load advantage on a stratified subset. A no-feedback second pass yields a small improvement over the original translations, and direct evaluator prompting reproduces the low-load advantage without a separate rewriting stage. These results show that compact feedback limits the quality loss from evaluator-guided refinement in the evaluated high-resource MT settings.
\end{abstract}

\begin{figure*}[htbp]
  \centering
  \includegraphics[width=0.98\linewidth]{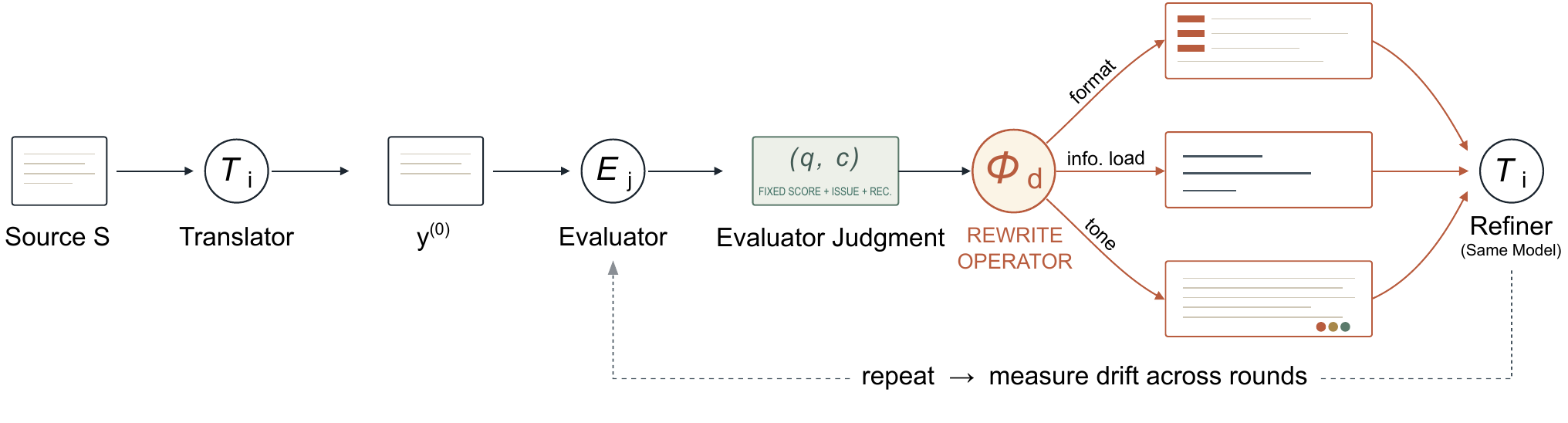}
  \caption{Controlled comparison for feedback presentation. The score, main issue, and recommendation are fixed, while the feedback is rewritten by tone, format, or information load before revision by the same translator.}
  \label{fig:fig1}
\end{figure*}

\section{Introduction}

When an LLM evaluator identifies an accuracy error in another model’s translation, the translator model is prompted to revise, and the resulting output may be unchanged, improved, or degraded. 
The same judgment can arrive as a terse instruction, a structured checklist, or a long explanation with qualifiers. These presentations may preserve the same score and core recommendation, yet an LLM reviser may respond to them differently. Feedback acts as an interface between evaluator judgment and revision.

This distinction matters because LLM refinement through iterative feedback is often brittle \cite{shinn2023reflexion, madaan2023selfrefineiterativerefinementselffeedback, chen2024selfdebug, kamoi-etal-2024-llms}. Across tasks, critique can help, do little, or degrade outputs \citep{pan-etal-2024-automatically, wang-etal-2025-cross, MAMM-Refine, huang2024large, tyen-etal-2024-llms, Jiang_Zhang_Weller_Weir_Van_Durme_Khashabi_2025}. The unresolved question is which part of the feedback signal a model uses when revising.

Existing work 
usually varies both the identity of the critic (i.e., the evaluator model used to generate feedback)
and the way feedback is expressed \citep{wang-etal-2025-cross, gou2024critic, paul-etal-2024-refiner, subramaniam2025multiagent}. As a result, it is difficult to separate \emph{who} critiques from \emph{how} the same core critique is presented.

We isolate this sensitivity with a controlled protocol. For each source sentence and evaluator comment, we keep the score, main issue, and recommendation fixed while changing the feedback's wording, 
format, or amount of exposed detail. The final suite tests two surface-form dimensions, tone and response format, plus one content-density dimension, information load. 
Because rewrites can alter the underlying critique, rewrite faithfulness and style compliance are checked explicitly. 

We therefore ask three questions. RQ1: Does changing feedback form or information load affect revision quality when the evaluator's score, main issue, and recommendation are held fixed? 
RQ2: How \textbf{robust} 
are the observed effects across metrics and validation checks? RQ3: Does direct evaluator prompting reproduce the low-load advantage without a separate rewriting stage? We use evaluator identity as a baseline, with feedback form and density as the main variables of interest. Analyses of tone, response format, and a second refinement round are reported in Appendices~\ref{sec:surface_form} and~\ref{sec:two_round}.

\paragraph{Contributions.}
We make three contributions. First, we introduce an MT refinement evaluation protocol and dataset artifacts that separate critic identity from feedback form and supporting detail across seven models and six translation directions. Second, we show that low-load feedback limits quality degradation relative to high-load feedback, with support from xCOMET and a MetricX-24 subset evaluation. Third, we evaluate a no-feedback second pass and direct evaluator prompting to examine revision without evaluator feedback and to test compact feedback without an additional rewriting stage.


\section{Related Work}



\paragraph{LLM Judging and Refinement} 
Self-refinement and self-correction studies report both gains and failures \citep{kamoi-etal-2024-llms, pan-etal-2024-automatically, huang2024large, tyen-etal-2024-llms, Jiang_Zhang_Weller_Weir_Van_Durme_Khashabi_2025}. Later variants add reranking, tools, confidence estimates, or reasoning diversity \citep{yan2023refining, yao2023react, wang2023selfconsistency, yao2023tree}. Cross-refinement moves critique to a different model and often improves over pure self-correction \citep{wang-etal-2025-cross, gou2024critic, paul-etal-2024-refiner}. Multi-agent and ensemble pipelines add debate, adjudication, or consensus \citep{irving2018ai,subramaniam2025multiagent,rawat2025pre,omar2024refining,motwani2024malt,chen2024magicore}. These systems show that refinement is conditional and that critic identity can matter. They also tend to vary critic identity, critique content, roles, instructions, evidence, and wording together, leaving the feedback interface itself uncontrolled.

\paragraph{Varying the Presentation.}
Prompt-steerability and evaluator-bias studies show that wording and presentation can steer model judgments \citep{miehling-etal-2025-prompt-steerability,wu-aji-2025-style-over-substance,fu-liu-2025-how-reliable,hong-etal-2026-beyond}. Related feedback-design work also varies instruction style or feedback phrasing \citep{zhou2023leasttomost, 2024.EDM-short-papers.49, leiter-eger-2024-prexme}. 
These studies motivate the interface view, but their outcomes usually concern judgments, preferences, or aggregate system performance. Our setting asks a more systematic and controlled question about revision: 
if the evaluator score, main issue, and recommendation are fixed, can feedback form or the amount of detail exposed to the reviser 
still change the revised translation?

We pair controlled rewrites with explicit faithfulness and style-compliance checks \citep{2024.EDM-short-papers.49}. This lets us test feedback presentation as a controllable variable under semantic constraints.


\section{A Controlled Feedback Framework for LLM Refinement}

This section defines the paper's central comparison: critic identity varies at the evaluator stage, while feedback form and exposed detail vary through controlled rewrites of the same evaluator judgment.
Surface-form variants change how the critique looks; information-load variants change how much of the critique the reviser sees. Both preserve the score, main issue, and recommendation.

Table~\ref{table:feedback_dimensions} summarizes the three dimensions, and Table~\ref{table:feedback_example} gives a concrete example from the rewrite logs.

\begin{table}[t]
\centering
\small
\setlength{\tabcolsep}{3pt}
\renewcommand{\arraystretch}{1.08}
\begin{tabular}{p{0.28\linewidth}p{0.62\linewidth}}
\toprule
\textbf{Factor} & \textbf{Values and change} \\
\midrule
Tone & Positive, neutral, negative. Changes evaluative wording. \\
Response format & Structured, free-text. Changes organization. \\
Information load & Low load, high load. Changes supporting detail. \\
\bottomrule
\end{tabular}
\caption{Controlled feedback dimensions. Tone and response format are surface-form manipulations; information load changes the amount of supporting detail shown to the reviser.}
\label{table:feedback_dimensions}
\end{table}


\subsection{Translator and Evaluator Refinement Framework}
As shown in Figure~\ref{fig:fig1}, our setup has two roles: a translator and an evaluator, which may be the same or different underlying models.
We use machine translation (MT) because small lexical or structural revisions can materially change quality, and sentence-level automatic metrics make large controlled comparisons feasible. MT is also a common testbed for iterative LLM refinement \citep{madaan2023selfrefineiterativerefinementselffeedback, pan-etal-2024-automatically, kocmi-federmann-2023-large}.

In each round, the translator receives a source sentence and produces a translation. An evaluator is prompted to score the translation on a 0 to 100 scale using a prompt adapted from GEMBA\footnote{Prompts for translator and evaluator roles are provided in the Appendix.} \citep{kocmi-federmann-2023-large}. The evaluator sees only the source sentence and the candidate translation; no human reference is provided. We retain the full generated evaluator output as evaluation text and parse the numerical score separately. When the output contains explanatory text in addition to the requested score, that text is retained as feedback. The original translator receives the parsed score and retained evaluation text and produces a single revised translation. The evaluator may be identical to the translator (self feedback) or a different model (cross-model feedback). In controlled-rewriting conditions, a separate model rewrites the evaluator's evaluation text before it is sent back to the translator; the evaluator score and core evaluative content, namely the main issue and recommendation, are kept fixed.

For each comparison, the source, translator, evaluator, score, main issue, and recommendation remain fixed, while the displayed feedback variant changes. Surface-form variants change tone or organization. Information-load variants preserve the core judgment while changing how much supporting detail the reviser sees. 
Low-load feedback preserves the score, main issue, and recommendation while reducing supporting detail. The original translator then revises from the same initial translation under each feedback variant.

\subsection{Controlled Feedback Dimensions}
The experiment suite directly controls three observable properties of the evaluator comment. The target is preservation of \emph{core evaluative content}: the score, main issue, and main recommendation. Tone and format are surface-form manipulations. Information load is a content-density manipulation: low-load feedback deliberately changes how much supporting detail is exposed to the reviser.


The main rewriting experiment uses GPT-OSS-20B, with prompts that instruct it to preserve the score, named entities, main issue, and recommendation. Additional experiments with direct structured or neutral feedback, near-identity rewriting, and Llama-3.1-8B as a second paraphraser are reported in Section~\ref{sec:valid_checks}. Section~\ref{sec:direct_prompting} evaluates direct low-load and high-load feedback generation. We assess content preservation in Section~\ref{sec:valid_checks}.

\paragraph{Tone.}
Tone prompts make the same core critique sound more positive, neutral, or negative while preserving the core issue and recommendation. We test these rewrites to measure whether sentiment-laden language helps or distracts the translator during revision \citep{Van-DijkxFeedback}.

\paragraph{Response Format.}
Format controls whether the same feedback is presented as a structured template or as a single free-text paragraph. This isolates the contribution of explicit organization without changing the underlying criticism.

\paragraph{Information Load.}
This control varies information load: low-load rewrites keep the score, core issue, and direct recommendation while removing secondary detail and qualifiers where possible, whereas high-load rewrites preserve the fuller detail 
and qualification structure of the original comment \citep{paas_renkl_sweller_2003, cese_cognitive_load_report_2017}. Length changes as a surface consequence: low-load rewrites average roughly 0.45$\times$ the original character length, while high-load rewrites remain near parity (Appendix~\ref{app:valid_check}).

The intended construct is the amount of supporting detail made available to the revising model.

\begin{table*}[t]
\centering
\footnotesize
\setlength{\tabcolsep}{4pt}
\renewcommand{\arraystretch}{1.15}
\begin{tabular}{p{0.13\textwidth}p{0.81\textwidth}}
\toprule
\textbf{Variant} & \textbf{Feedback shown to the reviser} \\
\midrule
\textbf{Raw} &
The translation captures the main idea of the German sentence but has issues with accuracy and naturalness. The message about heart disease as a leading cause of death is conveyed. However, ``more than any other single factor'' misrenders ``als alles andere zusammen,'' which means ``than all others combined.'' The last sentence is also off: the German says heart disease is preventable for almost all people, while ``almost preventable in everyone'' changes the emphasis. A score around 75 reflects the general idea with inaccuracies and awkward phrasing. \\
\midrule
\textbf{Low load} &
Score: 75/100. Main issue: two meanings are inaccurate, ``than all others combined'' and ``preventable for almost all people.'' Recommendation: revise those phrases for accuracy and naturalness. \\
\midrule
\textbf{High load} &
The translation captures the main idea, but several accuracy and naturalness issues remain. The opening correctly conveys that heart disease is a leading cause of death worldwide. The comparative phrase weakens the source meaning: ``als alles andere zusammen'' means ``than all others combined''; the translation gives ``more than any other single factor.'' The final sentence also changes the emphasis. The German says heart disease is preventable for almost all people; ``almost preventable in everyone'' sounds awkward and suggests a different meaning. A score around 75 is appropriate because the message is understandable, but these two places need revision. \\
\midrule
\textbf{Structured} &
\textbf{Overall:} acceptable with notable inaccuracies.\newline
\textbf{Key reasons:} the core message is correct; ``more than any other single factor'' weakens ``than all others combined''; ``almost preventable in everyone'' changes the final sentence.\newline
\textbf{Recommendation:} revise the comparative phrase and final sentence.\newline
\textbf{Rating:} 75/100. \\
\bottomrule
\end{tabular}
\caption{Feedback example for one German-to-English item with a 75/100 evaluator score. The table lightly shortens logged feedback for space. All rows keep the same score, main issue, and recommendation, while varying how the critique is organized and how much supporting detail is included. Bold labels in the structured variant reflect the labeled-field format; other variants are shown without added formatting.}
\label{table:feedback_example}
\end{table*}

\section{Experimental Setup}

\subsection{Data Collection and Curation}
We use parliamentary text to obtain grammatical, naturally occurring sentences while reducing obvious overlap with standard MT benchmarks. We curate 1,000 English and German sentences from the Hansard corpus\footnote{\url{https://hansard.parliament.uk/}; UK Parliament content is available under the Open Parliament Licence v3.0 at \url{https://www.parliament.uk/site-information/copyright-parliament/open-parliament-licence/}.} and the official records of the Deutscher Bundestag\footnote{\url{https://www.bundestag.de/dokumente/protokolle/amtlicheprotokolle}; the Bundestag provides plenary records through its Open Data service at \url{https://www.bundestag.de/services/opendata}, while general website content remains subject to the Bundestag legal notice at \url{https://www.bundestag.de/en/service/legal_notice}.}, respectively. These sources provide formal but naturally occurring text suitable for translation evaluation. 

We sample 1,000 English sentences and 1,000 German sentences after sentence segmentation, language filtering, removal of obvious boilerplate, and exact-duplicate filtering in the final sentence list. We then run each translator on six directions: English$\rightarrow$German/Spanish/Chinese and German$\rightarrow$English/Spanish/Chinese. These model outputs serve as refinement objects and are distinct from human references. Unlike widely used MT benchmarks such as WMT, these parliamentary materials were curated from official record pages accessed in November 2024 and kept separate from standard MT benchmark releases. This avoids standard benchmark test-set reuse; exposure to public parliamentary records in model pretraining remains possible. We avoid standard benchmark test sets but do not claim source-text novelty.
The resulting benchmark-style artifact contains 42,000 initial translations (7 translator models 
$\times$ 6 directions $\times$ 1,000 sentences) and 294,000 evaluator feedback instances. The rewriting experiment uses a stratified subset of 29,400 comments, with 100 source items for each combination of translator, evaluator, and translation direction (7$\times$7$\times$6). All controlled rewrite families are generated from this same 29,400-comment base set. This gives equal coverage of all combinations while keeping rewriting, refinement, and metric evaluation computationally feasible. Applying the same design to all 294,000 evaluator comments would increase this workload by approximately a factor of ten.

We will release the sentence lists or source identifiers, generation scripts, prompts, translations, evaluator outputs, paraphrased feedback variants, and scoring scripts where redistribution is permitted; for any source text with unresolved redistribution terms, we will release URLs/IDs and derived artifacts, while restricted text stays unreleased.

\subsection{Model Selection}
We use seven open or openly available instruction-tuned models spanning 7B to 27B-class systems: LLaMA, Qwen, Mistral, Gemma, and GPT-OSS-20B.\footnote{In preliminary experiments, we also evaluated smaller models but found that models with approximately 3B parameters often struggled to follow refinement instructions reliably, introducing substantial noise that could confound the analysis. We therefore focus on models with at least 7B parameters to ensure sufficient instruction-following and refinement capability.} This gives a heterogeneous translator/evaluator matrix for multi-model comparison. The main matrix excludes frontier API-only evaluators such as GPT-4 or Claude because the goal here is a repeatable, large-scale benchmark. Full checkpoint names are provided in Table~\ref{table:model_selection}.

\subsection{Meta-Evaluation with xCOMET}
We evaluate metric-predicted refinement quality using xCOMET-XL,\footnote{\url{https://huggingface.co/Unbabel/XCOMET-XL}} an automatic MT quality metric where higher scores indicate better predicted translation quality \citep{guerreiro-etal-2024-xcomet}. We run xCOMET in reference-free mode, using only the source sentence and the candidate translation. xCOMET builds on the COMET family of metrics, which has demonstrated high correlation with expert human MQM annotations across recent benchmark evaluations \citep{kocmi2021ship}. It is a useful automatic proxy for sentence-level translation quality, but it remains one metric family. Prior work also shows that language-model-based evaluation can exhibit systematic bias, particularly in reference-free settings \citep{liu-etal-2024-llms-narcissistic}. Throughout the paper, we report xCOMET multiplied by 100 for readability. The reported findings on tone, response format, and two-round score changes are specific to xCOMET. The low-load versus high-load comparison is additionally evaluated with MetricX-24.
We use MetricX-24-Hybrid in reference-free mode \citep{juraska-etal-2024-metricx}. Both xCOMET and MetricX-24 are learned metrics, so agreement between them provides additional automatic-evaluation evidence.

\subsection{Controlled Feedback Rewriting}
The base-refinement condition uses the evaluators' original evaluation text. We paraphrase this 29,400-comment subset with GPT-OSS-20B while preserving each comment's score, main issue, and main recommendation. The paraphraser only rewrites the evaluator's existing comment and leaves the score unchanged. To ensure the validity of the rewrite, the evaluation text is checked through the 84-pair rewrite validity sample in Section \ref{sec:valid_checks} and Appendix \ref{app:valid_check}.

The final analysis reports three prompt families: tone (positive, neutral, negative), response format (structured, free-text), and information load (low/high-load variants).\footnote{The main rewriting subset contains 29,400 evaluator comments. Tone has three rewrite values, while response format and information load have two each.}

\paragraph{Rewriting prompt.} The rewriting model receives the evaluator output and a requested presentation value while being instructed to preserve the score, factual content, named entities, main issue, and recommendation. The exact prompt used in the main rewriting experiment is shown below.
\begin{tcolorbox}[colback=gray!10, colframe=gray!160, left=2pt, right=2pt, title=Prompt - Controlled Feedback Rewriting]
\small
Rewrite the evaluator feedback below to match the requested presentation style while preserving the underlying critique.

\medskip
\textbf{Requirements.}
Preserve the original score, factual content, named entities, language, main issue, and recommendation. Do not add new facts or remove key conclusions. Adjust only the requested presentation dimension. Output only the rewritten evaluation.

\medskip
\textbf{Input feedback.}
``\{\textit{evaluation\_text}\}''

\medskip
\textbf{Target dimension values.}
Tone: positive / neutral / negative. Response format: structured / free\_text. Information load: low / high.

\medskip
\textbf{Style guidance.}
Structured uses labeled fields; free-text uses one cohesive paragraph. Low load keeps only decision-critical points; high load keeps relevant details and qualifiers. Tone changes wording emphasis while preserving the same core critique.

\medskip
\textbf{Rewritten evaluation:}
\end{tcolorbox}

In every condition, the original translator performs the final revision, keeping the refiner fixed within each comparison. The remaining variation comes from evaluator choice and the feedback shown to the translator. For load conditions, this includes the amount of supporting detail; for all rewrite conditions, paraphraser artifacts remain a possible confound. The results 
therefore include direct-control, near-identity, and second-paraphraser checks. 

\paragraph{Implementation details.} The local generation scripts use the HuggingFace \texttt{transformers} text-generation pipeline with bfloat16 weights, \texttt{device\_map=auto}, and \texttt{return\_full\_text=False}. Generation is deterministic under the recorded scripts because no sampling parameters are enabled; translation, evaluator-feedback, refinement, and score-extraction calls use maximum new-token budgets of 400, 1000, 600, and 512 respectively. Batch sizes are 50 to 100 except for Gemma-family jobs, which use batch size 1. xCOMET is run through the COMET checkpoint interface with batch sizes of 50 to 100 on one GPU. Paired-bootstrap summaries use a fixed analysis seed of 23.

\section{Results}



We first report raw-feedback refinement and a no-feedback second pass (\S\ref{sec:raw_baseline}), followed by the information-load comparison (\S\ref{sec:info_density}), direct evaluator prompting (\S\ref{sec:direct_prompting}), and feedback preservation (\S\ref{sec:valid_checks}). Tone, response format, and two-round results are reported in Appendices~\ref{sec:surface_form} and~\ref{sec:two_round}.

\subsection{Raw Feedback and a No-Feedback Second Pass}
\label{sec:raw_baseline}
Base quality varies substantially by model and direction. This matters because weak translations leave room for correction while strong translations can be over-edited. German$\rightarrow$English is easiest on average, whereas Chinese-target directions are markedly harder; Table~\ref{tab:translation-results} reports the full base matrix.

\begin{table*}[t]
\centering
\small
\setlength{\tabcolsep}{5.5pt}
\renewcommand{\arraystretch}{1.12}

\begin{tabular}{p{2.7cm}ccccccc}
\toprule
\diagbox[width=8em,height=2.3\baselineskip]{Translator}{LP}
& En$\rightarrow$De
& En$\rightarrow$Es
& En$\rightarrow$Zh
& De$\rightarrow$En
& De$\rightarrow$Es
& De$\rightarrow$Zh
& Avg. \\
\midrule
Mistral-7B & 87.51 & 83.83 & 58.78 & 93.21 & 85.86 & 66.38 & 79.26 \\
Llama3.1-8B  & 91.07 & 86.65 & 68.76 & 94.34 & 88.58 & 71.95 & 83.56 \\
Qwen2.5-7B          & 86.66 & 85.87 & 72.06 & 92.74 & 83.95 & 68.65 & 81.66 \\
Qwen2.5-14B         & 90.39 & 88.02 & 73.64 & 93.10 & 84.55 & 72.86 & 83.76 \\
Gemma3-12B             & 93.36 & 89.97 & 73.78 & 92.60 & 89.47 & 69.88 & 84.84 \\
Gemma3-27B             & 93.91 & 90.16 & 75.40 & 92.70 & 89.49 & 71.91 & 85.60 \\
GPT-OSS-20B                & 92.78 & 88.91 & 72.41 & 94.90 & 89.68 & 74.62 & 85.55 \\
\midrule
Avg.                               & 90.81 & 87.63 & 70.69 & 93.37 & 87.37 & 70.89 & -- \\
\bottomrule
\end{tabular}
\caption{Base translation quality (xCOMET$\times 100$) by translator and language pair.}
\label{tab:translation-results}
\end{table*}


Table~\ref{table:base_refinement} summarizes the default refinement loop using the evaluators' original outputs. The overall mean falls from 83.46 for the original translations to 82.13 after one refinement round. Self feedback is the weakest one-round setting (81.40), while the cross-model critic selected retrospectively from the evaluated outputs, reaches 83.32. 

Brittleness varies by translator and direction. Weaker translators sometimes benefit from suitable cross-model critics, while stronger translators are often over-edited: even the retrospectively selected cross-model critic stays below the original score for Gemma3-12B, Gemma3-27B, Llama3.1-8B, and GPT-OSS-20B. Raw refinement helps German$\rightarrow$Chinese (+3.29 over the original average), hurts English$\rightarrow$Spanish (-5.01) and English$\rightarrow$German (-2.04). This conditional pattern motivates the central question: if critique frequently hurts, 
can controlling its form and density reduce the risk?

\begin{table*}[htbp]
  \centering
  \small
  \renewcommand{\arraystretch}{1.15}
  \begin{tabular}{lcccc}
  \toprule
  \textbf{Translator} & \textbf{Original} & \textbf{Self Feedback} & \textbf{Best Cross-Critic} & \textbf{Avg. over Critics} \\
  \midrule
  Mistral-7B & 79.26 & 75.16 & \textbf{80.53} & 78.47 \\
  Llama3.1-8B & 83.56 & 80.70 & 83.21 & 81.86 \\
  Qwen2.5-7B & 81.66 & \textbf{81.89} & \textbf{83.70} & \textbf{82.52} \\
  Qwen2.5-14B & 83.76 & 82.49 & \textbf{84.06} & 83.15 \\
  Gemma3-12B & 84.84 & 82.72 & 84.08 & 82.68 \\
  Gemma3-27B & 85.60 & 83.47 & 84.06 & 83.41 \\
  GPT-OSS-20B & 85.55 & 83.39 & 83.61 & 82.85 \\
  \midrule
  Avg. & 83.46 & 81.40 & 83.32 & 82.13 \\
  \bottomrule
  \end{tabular}
\caption{Mean refinement quality (xCOMET$\times100$) across six translation directions. Self feedback uses the same model as translator and evaluator. For each translator and translation direction, Best Cross-Critic selects the non-identical evaluator with the highest mean refinement score, then averages the selected scores across directions. Selection uses the evaluated outputs. Bold values exceed the original translation baseline.}
  \label{table:base_refinement}
\end{table*}

\paragraph{No-feedback second pass.}
The translator receives the source sentence and its original translation and produces a second version without evaluator feedback. On 2,700 examples, this condition improves xCOMET$\times100$ by 0.39 over the original translations (95\% CI [0.05, 0.70]).

\subsection{Low-Load Feedback Limits Quality Degradation}
\label{sec:info_density}
With evaluator judgment held fixed, different feedback variants lead to different revisions. Table~\ref{table:feedback_style} reports low-load and high-load results on the rewriting subset alongside full-matrix reference means for raw refinement and the original translations.

Low-load feedback outperforms high-load feedback in all 42 translator-direction settings, with a mean difference of 2.63 xCOMET$\times100$ (95\% CI [2.26, 3.05]). On the 29,400-record rewriting subset, the mean scores are 83.30 for low-load feedback and 80.67 for high-load feedback. For reference, the full-matrix means are 83.46 for the original translations and 82.13 for raw refinement. Low-load feedback therefore limits the quality loss relative to high-load feedback.

The MetricX-24 subset evaluation shows the same low-load advantage. On a stratified 1,008-example subset, we sample 24 feedback instances from each of the 42 translator-direction settings, with evaluator identities represented within each setting; low-load and high-load feedback use the same 1,008 common keys. After converting MetricX-24 predictions to higher-is-better scores, low-load feedback exceeds high-load feedback by 0.54 points (95\% bootstrap CI [0.32, 0.76]).\footnote{We report 25 minus the MetricX-24 raw prediction so that higher values indicate better predicted quality. This transformation reverses the direction of the original scale while preserving pairwise difference magnitudes.}

\begin{table}[t]
\centering
\small
\setlength{\tabcolsep}{6pt}
\begin{tabular}{lc}
\toprule
\textbf{Condition} & \textbf{xCOMET$\times100$} \\
\midrule
Original translation (full matrix) & 83.46 \\
Raw refinement (full matrix) & 82.13 \\
Low-load feedback & 83.30 \\
High-load feedback & 80.67 \\
\bottomrule
\end{tabular}
\caption{Mean xCOMET$\times100$ scores. Original translation and raw refinement are full-matrix reference means; low-load and high-load feedback are means on the 29,400-record rewriting subset.}
\label{table:feedback_style}
\end{table}


\subsection{Direct Evaluator Prompting}
\label{sec:direct_prompting}

We prompt evaluators to generate low-load or high-load feedback directly, without a separate rewriting stage or an injected score line. On a 5,400-example subset, the mean xCOMET$\times100$ scores are 68.42 for low-load feedback and 67.27 for high-load feedback, a difference of 1.15 (95\% CI [0.43, 1.85]). This subset has a different model and direction composition from the main rewriting experiment, so we interpret the difference within this subset. The evaluator can produce different judgments under the two prompts; this experiment evaluates direct feedback generation.

\subsection{Feedback Preservation and Validation}
\label{sec:valid_checks}

\paragraph{Faithfulness triage.}
We run a heuristic pass over 205,800 rewrite rows and flag cases where explicit scores, numeric spans, or quoted fragments appear to have changed or disappeared. Because the raw paraphraser logs often contain wrapper text and formatting artifacts, this stage is deliberately conservative and is used to prioritize follow-up inspection while avoiding a precise preservation-rate claim. It produces a stratified 84-pair annotation sheet covering all reported rewrite families. We then run a judge-based check over that full sample with Qwen2.5-14B and Gemma3-27B as independent judges, followed by third-judge adjudication on disagreement cases. The label set records whether the rewrite preserves the original issue, severity, and recommendation, whether it introduces a new unsupported issue, and whether it removes important detail. Adjudicated majority labels indicate that the rewrite preserves the original issue in 81/84 cases, preserves the recommendation in 81/84 cases, and introduces no new unsupported issue in 80/84 cases; a conservative strict-preservation criterion holds in 70/84 cases. By value, the weakest family remains negative tone (7/12 strict-preserved), while structured and low-load rewrites more often preserve the core issue than exact secondary detail. The scaled check therefore strengthens the methodological claim without licensing a statement of perfect semantic invariance.

\paragraph{Paraphraser artifacts.}
Concern: the paraphraser might create the apparent presentation effect. Check: we prompt the evaluator itself to produce structured or neutral feedback (direct control), then rerun the same subset with Llama-3.1-8B as a second paraphraser. Result: the direct-control condition partially reproduces the same ordering (structured 76.88, neutral 76.30, gap +0.57; 95\% CI [0.05, 1.08]), and the second paraphraser preserves the same qualitative picture. A single-paraphraser explanation becomes less likely.

\paragraph{Generic rewriting effects.}
A near-identity rewrite keeps feedback close to its original wording. Its mean xCOMET$\times100$ score is 1.87 points below raw refinement on the same examples (95\% CI [$-2.27$, $-1.45$]). Appendix~\ref{sec:two_round} reports its scores across two refinement rounds.

\paragraph{Illustrative validation patterns.}
The failure modes in the scaled check are also informative. Negative-tone failures rarely invent a completely unrelated critique; they more often compress a largely positive or mixed assessment into language that sounds globally harsher, shifting severity and occasionally implying a new unsupported issue. Structured and low-load rewrites show a different pattern: they usually retain the main recommendation, but sometimes compress away supporting evidence, quoted phrases, or numeric detail that justified the original score. These patterns show that the requested presentation changes are usually realized, although severity and supporting detail can change during rewriting.

\paragraph{Representative cases.}
The sampled cases reflect both sides of this pattern. One preserved example comes from a Qwen$\rightarrow$Chinese rewrite where a free-text paraphrase keeps the original score, core issue, and wording adjustments almost verbatim while removing only superficial redundancy. One risky negative-tone example comes from a German$\rightarrow$English case where the original comment assigns 95/100 with only a minor stylistic qualification, then the rewrite compresses that assessment into a more globally unfavorable judgment. One risky structured example comes from an English$\rightarrow$German case where the main recommendation survives, although the rewrite drops quoted evidence and numeric detail that justified the score. These cases match the aggregate validation result: the main issue is usually preserved, while severity and supporting detail are the first things to drift.

\paragraph{Style-compliance check.}
We also measure whether requested surface-form controls were actually realized. Response-format control is strongest: structured rewrites almost always exhibit list-like or labeled organization, while free-text rewrites remain single-paragraph prose. Information-load control is also visible on the surface, with low-load rewrites averaging roughly 0.45$\times$ the original character length and high-load rewrites staying near parity. Tone control is noticeably noisier under simple lexical heuristics, especially for negative and neutral prompts. The clearest evidence is therefore for steerable structure and information load, while tone should be interpreted more cautiously unless stronger validation is added.

\paragraph{Feedback length.}
Within each load condition, feedback length has little association with xCOMET scores ($|\rho|<0.03$). A mixed-effects regression over 58,800 low-load and high-load rows models xCOMET as a function of feedback condition and log feedback character count, with translator random intercepts and language-pair variance components. The high-load coefficient is $\beta=-4.02$ (95\% Wald CI [$-4.42$, $-3.62$], $p<0.001$).






\section{Discussion}

\paragraph{Feedback as an interface.}
The results show that feedback presentation affects how LLMs use evaluator judgments during revision. Preserving the score and core recommendation still permits different revision behavior depending on how the same critique is organized or compressed. A refinement pipeline should therefore treat critic identity, critique content, and presentation form as separate design choices.

\paragraph{Connecting to feedback-design work.}
The older feedback-design vocabulary remains useful as an interpretation layer, although the controlled experiments here measure form and exposed detail without measuring psychological constructs. Compact feedback increases directness by removing secondary explanations; structured feedback changes organization; neutral tone changes framing but with weaker and less robust evidence.

\paragraph{Possible explanations.}
Additional supporting detail may direct attention to several revision targets. Compact feedback may also resemble the direct instructions used to train instruction-following models. Distinguishing these explanations requires targeted experiments, such as equal-length feedback variants or score-only, issue-only, and recommendation-only baselines.

\paragraph{Practical implications.}
In pipelines that already use evaluator-guided refinement, low-load feedback can be generated directly when evaluator prompts can be changed. A separate rewriting stage remains useful when evaluator feedback is fixed or supplied externally. The direct-prompt experiment supports this application on its evaluation subset.

\section{Conclusion}
Across the evaluated models and translation directions, low-load feedback produces less quality degradation than high-load feedback under xCOMET and shows the same comparative advantage under MetricX-24. Its subset mean is close to the full-matrix original-translation reference mean. The no-feedback second-pass experiment provides a baseline for revision without evaluator feedback, and direct evaluator prompting reproduces the low-load advantage without a separate rewriting stage. These findings characterize feedback use in the evaluated high-resource MT settings.

\section*{Limitations}
\paragraph{Claim scope.}
The main finding concerns the relative effects of low-load and high-load feedback in the evaluated models and translation directions. Low-load feedback yields higher scores than high-load feedback, and its subset mean is close to the full-matrix original-translation reference mean. Tone, response format, and two-round results are specific to xCOMET. The two-round comparison describes score changes from different round-one starting points on the English$\rightarrow$German and German$\rightarrow$Chinese subset.

\paragraph{Load construct.}
Information load is implemented by changing exposed supporting detail while preserving the score, main issue, and recommendation. The feedback-length analysis leaves the contributions of supporting detail, feedback length, and resemblance to instruction-style prompts unresolved. The experiments also lack low-load and high-load variants of equal length, as well as score-only, issue-only, and recommendation-only baselines. Multi-turn interaction and richer conversational feedback strategies remain open avenues.

\paragraph{Rewrite validity.}
Most rewrite experiments use GPT-OSS-20B as the paraphraser, and GPT-OSS-20B also appears in the translator/evaluator matrix. Direct-control, near-identity, and second-paraphraser checks reduce this concern, yet paraphraser-specific behavior remains possible. The faithfulness and style-compliance checks are sampled, so the strongest claims concern downstream behavior under controlled approximations; perfect semantic invariance is outside the supported scope.

\paragraph{Task and evaluation scope.}
The study is confined to high-resource machine translation directions and primarily to automated xCOMET evaluation, with MetricX-24 evaluated on a stratified subset. Human MQM-style judgments, additional independent metrics, lower-resource directions, and non-translation tasks would be needed before making broader quality or transfer claims. The parliamentary source text avoids standard MT benchmark releases, while exposure through public-record pretraining remains possible.

\section*{Ethical Considerations}
The study uses public parliamentary proceedings and model-generated translations, critiques, and revisions. The data may include names and political statements by public officials, so released artifacts will preserve source attribution and avoid suggesting endorsement by the source institutions. The work is a diagnostic study of refinement behavior outside deployment as a translation or political analysis system; generated translations and critiques may contain metric-preferred but human-unverified errors.

Release is governed by source terms. Hansard material is reused under the Open Parliament Licence, whereas Bundestag website content and Open Data materials are subject to the Bundestag's published terms. We will therefore release scripts, prompts, model-output artifacts, scoring outputs, and source IDs/URLs, and will redistribute raw source text only where the applicable terms permit it.




\bibliography{custom}


\clearpage
\appendix

\section{Appendix Roadmap}

\hyperref[app:model_selection]{Appendix~\ref*{app:model_selection}} summarizes checkpoint metadata, \hyperref[app:prompt_design]{Appendix~\ref*{app:prompt_design}} gives the active prompt families, \hyperref[app:controlled_dimensions]{Appendix~\ref*{app:controlled_dimensions}} records the controlled dimensions, \hyperref[app:valid_check]{Appendix~\ref*{app:valid_check}} gives representative validation cases, and \hyperref[app:base_refinements]{Appendix~\ref*{app:base_refinements}} visualizes the full translator and evaluator refinement grid. \hyperref[sec:surface_form]{Appendix~\ref*{sec:surface_form}} and \hyperref[sec:two_round]{Appendix~\ref*{sec:two_round}} report tone and response-format effects and two-round xCOMET results, respectively.


\section{Model Selection}
\label{app:model_selection}
Table~\ref{table:model_selection} summarizes the checkpoints used in the main matrix together with publicly stated cutoff information where it is available.

\begin{table*}[ht]
\hspace*{-.8cm}
\centering
\renewcommand{\arraystretch}{1.3}
\resizebox{.8\textwidth}{!}{
\begin{tabular}{p{2.5cm} p{5.8cm} p{5cm} p{3cm}}
\toprule
\textbf{Model Name} & \textbf{Name of Checkpoint} & \textbf{Training Data Cutoff} & \textbf{Before Nov.\ 2024?} \\
\midrule
Llama-3.1-8B  & {meta-llama/Llama-3.1-8B-Instruct} &  2023-12   & \checkmark \\
Qwen2.5-7B    & {Qwen/Qwen2.5-7B-Instruct} &         \makecell[l]{Not publicly disclosed\\(before 2024-09)}
          & \checkmark  \\
Qwen2.5-14B    & {Qwen/Qwen2.5-14B-Instruct} &        \makecell[l]{Not publicly disclosed\\(before 2024-09)}
           & \checkmark  \\
Mistral-7B    & {mistralai/Mistral-7B-Instruct-v0.3} &  \makecell[l]{Not publicly disclosed\\(before 2024-05)}
         & \checkmark  \\
Gemma3-12B    & {google/gemma-3-12b-it} &             2024-08   & \checkmark  \\
Gemma3-27B    & {google/gemma-3-27b-it} &             2024-08   & \checkmark  \\
GPT-OSS-20B    &  {openai/gpt-oss-20b}  &            2024-06
 & \checkmark   \\
\bottomrule
\end{tabular}}
\caption{Selected models used as both translators and evaluators.}
\label{table:model_selection}
\end{table*}



\section{Prompt Design}
\label{app:prompt_design}
This appendix documents the core prompt families used in the main rewriting experiment: translation, evaluation, refinement, and controlled feedback rewriting. Exact prompts for the no-feedback second-pass and direct evaluator-prompting follow-up experiments are included in the released experiment scripts.

\section{Controlled Dimensions}
\label{app:controlled_dimensions}
Table~\ref{table:attributes_aspects} summarizes the controllable feedback dimensions used in the final experiments.

\begin{table*}[t]
\centering
\small
\setlength{\tabcolsep}{4pt}
\renewcommand{\arraystretch}{1.15}
\begin{tabular}{p{0.17\textwidth}p{0.28\textwidth}p{0.22\textwidth}p{0.22\textwidth}}
\toprule
\textbf{Dimension} & \textbf{What changes} & \textbf{Values} & \textbf{Held fixed} \\
\midrule
Tone & Surface wording and sentiment & Positive, neutral, negative & Score, main issue, recommendation \\
Response format & Organization of the same critique & Structured, free-text & Score, main issue, recommendation \\
Information load & Amount of supporting detail shown to the reviser & Low load, high load & Score, main issue, recommendation \\
\bottomrule
\end{tabular}
\caption{Controlled feedback dimensions used in the final experiments. Tone and response format manipulate surface form; information load manipulates accessible detail while keeping the core judgment fixed.}
\label{table:attributes_aspects}
\end{table*}

\subsection{Prompts for Translation, Evaluation, and Refinement}
The boxes below show the English$\rightarrow$German instantiation used for illustration; the same templates are applied to the other directions by swapping language names and placeholders.

\hspace*{-.5cm}
\resizebox{.45\textwidth}{!}{
\begin{tcolorbox}[colback=gray!10, colframe=gray!160, left=2pt, title=Prompt - Translation]

Translate the following English sentence into German.  Please provide the translation only. Please be brief and do not include any notes or discussion. \\

English sentence: \{\textit{sentence}\} \\

German sentence: 
\end{tcolorbox}
}

\hspace*{-.5cm}
\resizebox{.45\textwidth}{!}{
\begin{tcolorbox}[colback=gray!10, colframe=gray!160, left=2pt, title=Prompt - Evaluation]
Score the following translation from English to German on a continuous scale from 0 to 100, where a score of zero means "no meaning preserved" and a score of one hundred means "perfect meaning and grammar".\\
                        
English source: "\{\textit{source\_seg}\}"\\

German translation: "\{\textit{target\_seg}\}"\\

Score:
\end{tcolorbox}
}

\paragraph{Evaluator-output handling.} The prompt above is the executed evaluator prompt and explicitly requests a numerical score. We retain the full generated output as the evaluation text and parse the score separately; explanatory text generated alongside the score is retained as feedback for refinement.

\hspace*{-.5cm}
\resizebox{.45\textwidth}{!}{
\begin{tcolorbox}[colback=gray!10, colframe=gray!160, left=2pt, title=Prompt - Refinement]
You previously provided the following translation from English to German:\\

Source Sentence: \{\textit{source\_seg}\}\\
Your Original Translation: \{\textit{target\_seg}\}\\

The translation was evaluated by a mentor who scored it \{\textit{mentor\_score}\} out of 100.\\

Feedback:\\
"\{\textit{mentor\_explanation}\}"\\

Task:\\
Revise it to improve meaning preservation, grammatical fluency, and accuracy as needed. If you think no changes are needed (based on the feedback), feel free to leave it unchanged.
\end{tcolorbox}
}



\section{Validation Checks for Controlled Rewrites}
\label{app:valid_check}
We report supporting analyses of rewrite generation, content preservation, style compliance, and feedback length. These analyses characterize how the rewriting stage behaves and where the requested presentation changes can alter secondary details.

\section{Results of Base Refinements}
\label{app:base_refinements}
\begin{figure*}[htbp]
  \centering
  \includegraphics[width=.97\linewidth]{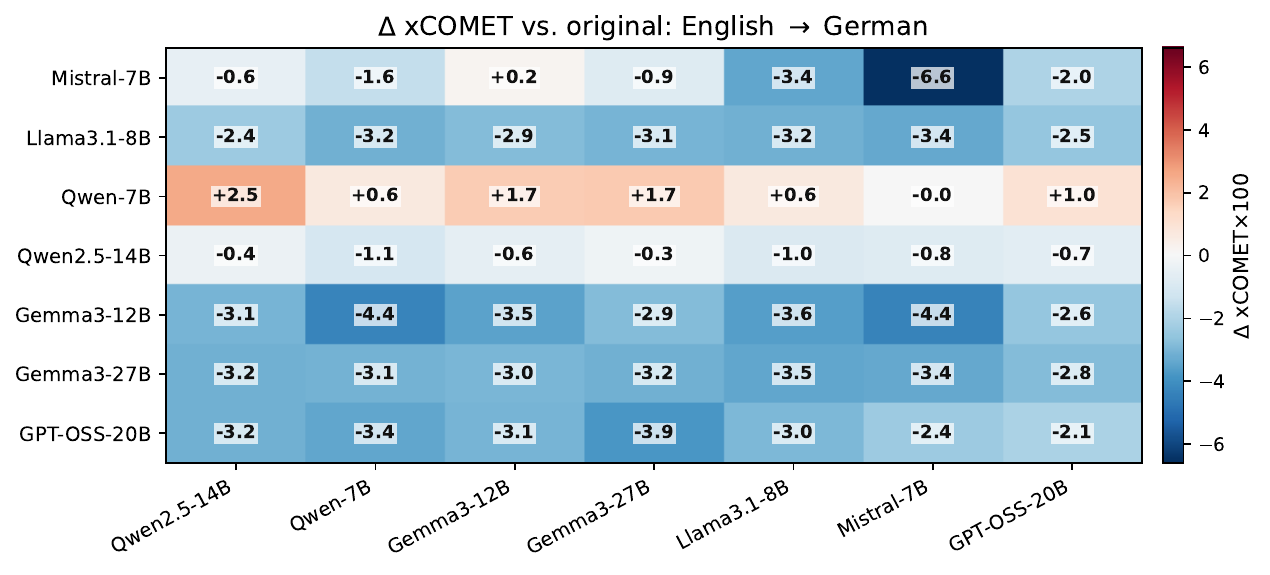}\par\vspace{0.4em}
  \includegraphics[width=.97\linewidth]{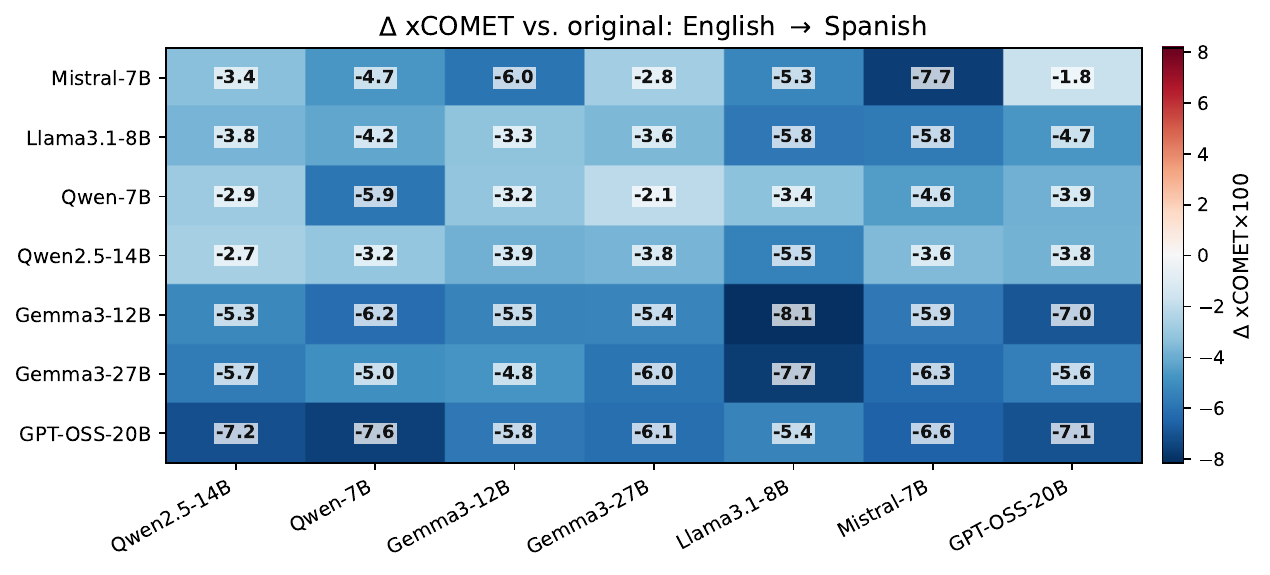}
  \caption{Appendix heatmaps for English-source directions. The top panel shows English$\rightarrow$German and the bottom panel shows English$\rightarrow$Spanish. Positive cells indicate that refinement helps relative to the original translation; negative cells indicate degradation.}
  \label{fig:appendix_heatmaps_en}
\end{figure*}

\begin{figure*}[htbp]
  \centering
  \includegraphics[width=.97\linewidth]{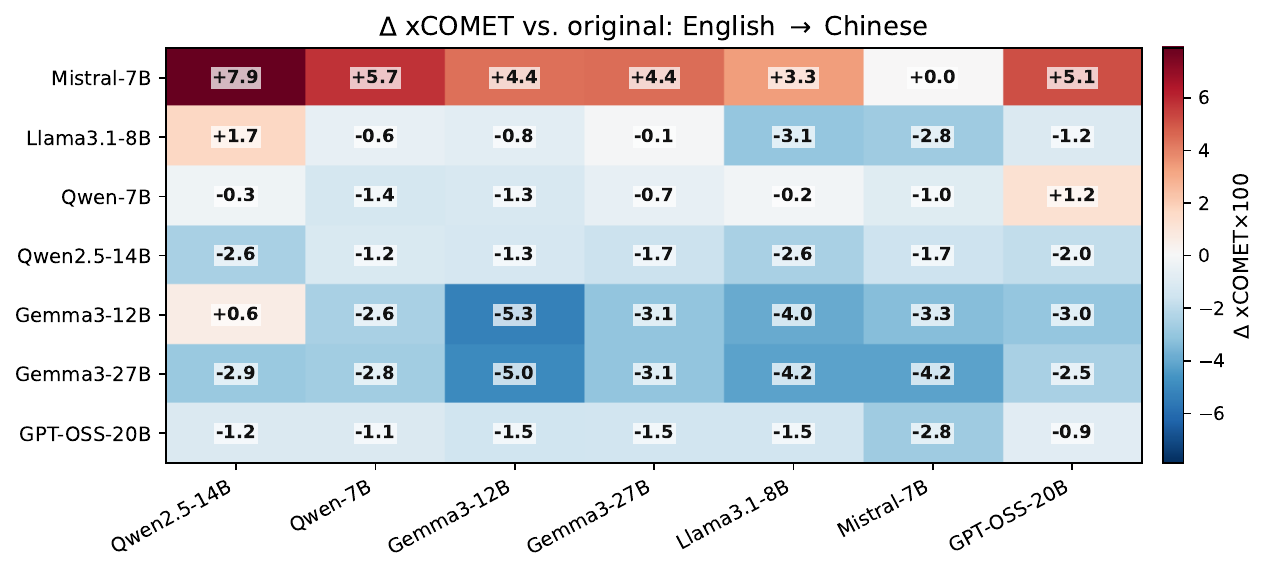}\par\vspace{0.4em}
  \includegraphics[width=.97\linewidth]{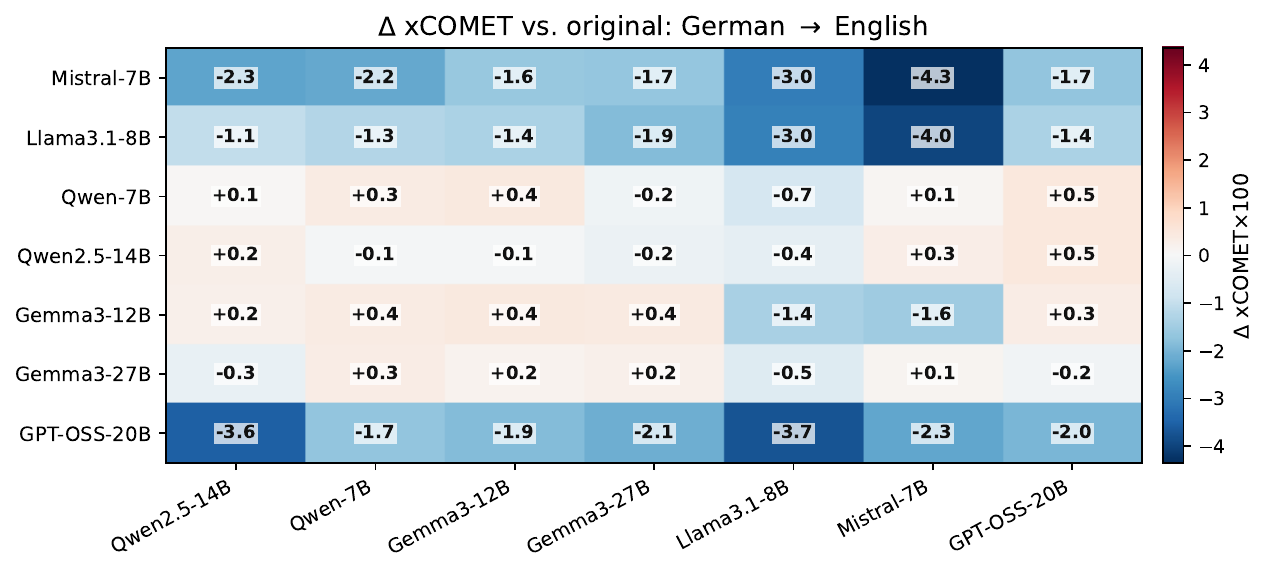}
  \caption{Appendix heatmaps for English$\rightarrow$Chinese (top) and German$\rightarrow$English (bottom). These panels contrast one harder Chinese-target setting with one easier English-target setting under the same translator and evaluator grid.}
  \label{fig:appendix_heatmaps_mix}
\end{figure*}

\begin{figure*}[htbp]
  \centering
  \includegraphics[width=.97\linewidth]{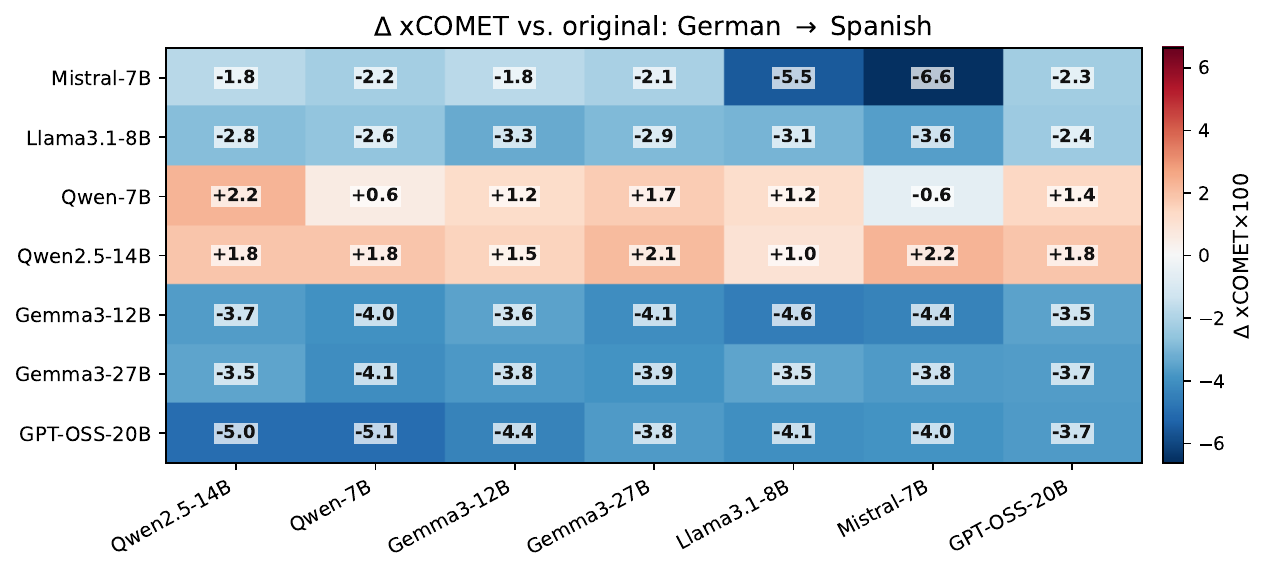}\par\vspace{0.4em}
  \includegraphics[width=.97\linewidth]{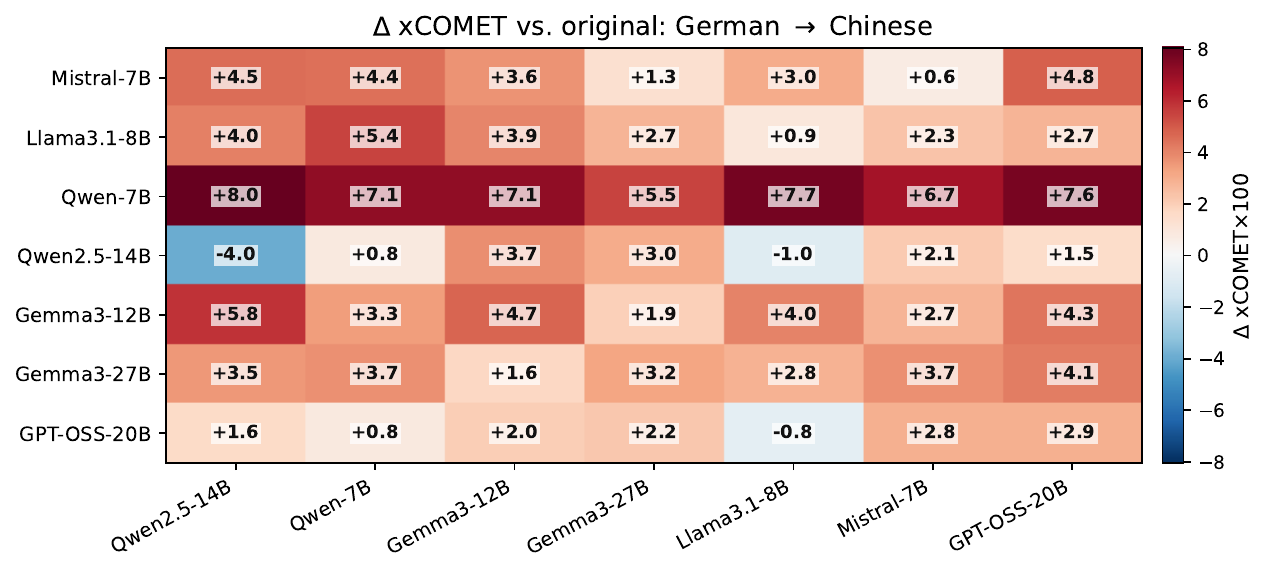}
  \caption{Appendix heatmaps for the remaining German-source directions: German$\rightarrow$Spanish (top) and German$\rightarrow$Chinese (bottom). Across all appendix panels, warmer colors indicate mean improvement and cooler colors indicate mean degradation relative to the original translation.}
  \label{fig:appendix_heatmaps_de}
\end{figure*}

Figures~\ref{fig:appendix_heatmaps_en} --- ~\ref{fig:appendix_heatmaps_de} report mean $\Delta$ xCOMET$\times 100$ relative to the original translation for each translator and evaluator pair. The diverging color scale is centered at zero, so positive cells indicate helpful refinement and negative cells indicate degradation.

\FloatBarrier
\section{Tone and Response Format under xCOMET}
\label{sec:surface_form}
Structured, free-text, and neutral rewrites score above raw refinement under xCOMET. The reported tone and format findings remain metric-dependent. Table~\ref{table:surface_form} reports these results.

\begin{table}[t]
\centering
\small
\setlength{\tabcolsep}{4pt}
\begin{tabular}{lcc}
\toprule
\textbf{Condition} & \textbf{Mean} & $\boldsymbol{\Delta}$ \textbf{vs. Raw} \\
\midrule
Raw refinement (reference) & 82.13 & 0.00 \\
Structured & 83.69 & +1.56 \\
Free-text & 83.43 & +1.30 \\
Neutral & 83.50 & +1.37 \\
Positive tone & 81.78 & -0.35 \\
Negative tone & 81.69 & -0.45 \\
\bottomrule
\end{tabular}
\caption{Tone and response-format results under xCOMET$\times100$ on the rewriting data. Raw refinement is shown as a full-matrix reference mean; differences are relative to that reference.}
\label{table:surface_form}
\end{table}

Raw feedback is generated without an explicit requirement on tone or organization. Neutral and free-text prompts impose additional presentation requirements, so their outputs can differ from raw feedback. The observed differences also include the effects of paraphrasing.

\begin{figure}[htbp]
  \centering
  \includegraphics[width=\linewidth]{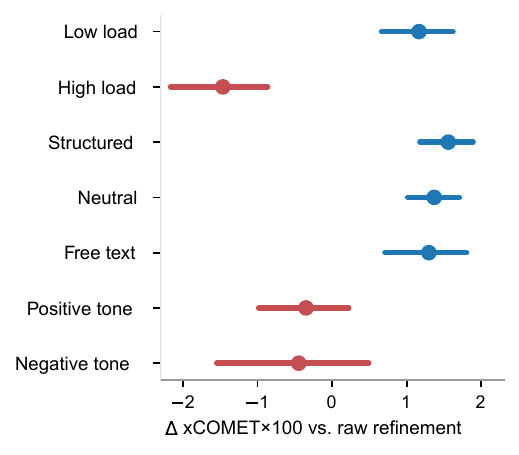}
\caption{Setting-level xCOMET$\times100$ differences relative to the raw-refinement reference across 42 translator-direction settings.}
  \label{fig:effect_sizes}
\end{figure}

Structured and neutral rewrites have xCOMET means 1.56 and 1.37 points above the full-matrix raw-refinement reference, respectively. The MetricX subset leaves structured feedback close to raw ($\Delta=-0.04$, 95\% CI [-0.15, 0.07]), so these surface-form results remain metric-dependent. Figure~\ref{fig:effect_sizes} summarizes setting-level deltas relative to raw refinement.

Tone is weaker and noisier: both positive and negative rewrites underperform raw refinement, and Appendix~\ref{app:valid_check} shows that negative rewrites have the lowest preservation rate among all rewrite families.



\FloatBarrier
\section{Two-Round xCOMET Results}
\label{sec:two_round}
We examine xCOMET score changes across two refinement rounds. We use a two-round subset with three translators (Mistral-7B, Qwen2.5-14B, Gemma3-27B), two evaluators, and two directions (English$\rightarrow$German, German$\rightarrow$Chinese), using 150 sentences each (1,800 examples per condition). In each case, the same evaluator re-critiques the round-one revision and the translator revises again.

Raw feedback scores 81.95 in round one and 78.79 in round two, a change of $-3.16$ (95\% CI [$-3.99$, $-2.37$]). The corresponding changes are $-0.54$ for near-identity feedback and $-0.94$ for low-load feedback. At round two, the differences relative to raw feedback are +0.11 for near-identity feedback and $-0.30$ for low-load feedback.

\begin{figure}[htbp]
  \centering
  \includegraphics[width=0.93\linewidth]{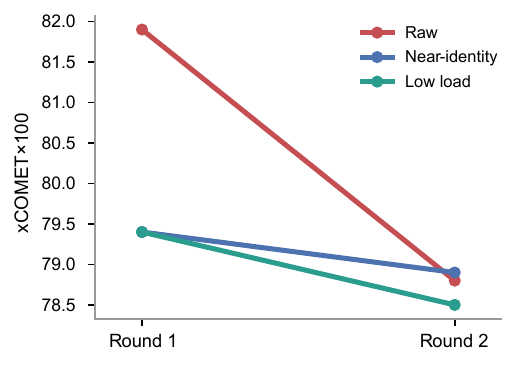}
\caption{Mean xCOMET$\times100$ scores across two refinement rounds on the English$\rightarrow$German and German$\rightarrow$Chinese subset. Raw feedback has a higher round-one score, and the conditions reach similar scores in round two. Table~\ref{table:two_round_stability} reports results by translator.}
  \label{fig:two_round_drift}
\end{figure}

This comparison describes round-to-round score changes from different round-one starting points. The final scores are similar across conditions. Table~\ref{table:two_round_stability} reports results by translator, and Figure~\ref{fig:two_round_drift} shows the aggregate scores.

\begin{table}[ht]
\centering
\scriptsize
\renewcommand{\arraystretch}{1.04}
\setlength{\tabcolsep}{2pt}
\begin{tabular}{lccc}
\toprule
\textbf{Translator} & \textbf{Raw} & \textbf{Near-id.} & \textbf{Low load} \\
\midrule
Mistral-7B & 79.7$\to$74.2 (-5.5) & 75.5$\to$74.6 (-0.9) & 77.1$\to$75.4 (-1.7) \\
Qwen2.5-14B & 82.1$\to$78.9 (-3.2) & 78.5$\to$79.3 (+0.8) & 78.1$\to$76.9 (-1.2) \\
Gemma3-27B & 84.0$\to$83.2 (-0.8) & 84.3$\to$82.8 (-1.5) & 83.1$\to$83.1 (+0.1) \\
\midrule
All translators & 81.9$\to$78.8 (-3.2) & 79.4$\to$78.9 (-0.5) & 79.4$\to$78.5 (-0.9) \\
\bottomrule
\end{tabular}
\caption{Two-round xCOMET$\times100$ results on the English$\rightarrow$German and German$\rightarrow$Chinese subset. Cells show round-one and round-two scores; parentheses show the change from round one to round two.}
\label{table:two_round_stability}
\end{table}

\end{document}